\documentclass[sn-mathphys]{sn-jnl}% Math and Physical Sciences Reference Style
\jyear{2022}%

\theoremstyle{thmstyleone}%
\theoremstyle{thmstyletwo}%

\usepackage{multirow}
\usepackage{bm}
\usepackage{ifthen}
\usepackage{tabularx}
\usepackage{xcolor}
\usepackage{textcomp}
\usepackage{stfloats}
\usepackage{url}
\usepackage{verbatim}
\usepackage{graphicx}
\usepackage{amsmath,amsfonts}
\usepackage{array}
\usepackage{lineno,hyperref}

\theoremstyle{thmstylethree}%

\begin{document}

\title[Pose-varied Face Restoration]{Enhancing Quality of Pose-varied Face Restoration with Local Weak Feature Sensing and GAN Prior}

%%=============================================================%%
%% Prefix	-> \pfx{Dr}
%% GivenName	-> \fnm{Joergen W.}
%% Particle	-> \spfx{van der} -> surname prefix
%% FamilyName	-> \sur{Ploeg}
%% Suffix	-> \sfx{IV}
%% NatureName	-> \tanm{Poet Laureate} -> Title after name
%% Degrees	-> \dgr{MSc, PhD}
%% \author*[1,2]{\pfx{Dr} \fnm{Joergen W.} \spfx{van der} \sur{Ploeg} \sfx{IV} \tanm{Poet Laureate} 
%%                 \dgr{MSc, PhD}}\email{iauthor@gmail.com}
%%=============================================================%%
\author[1]{\sur{Kai Hu}} %\email{kaihu@tju.edu.cn}

\author[1]{\sur{Yu Liu}} %\email{liuyu@tju.edu.cn}

\author[1]{\sur{Renhe Liu}} %\email{liurenhe@tju.edu.cn}
%\equalcont{These authors contributed equally to this work.}
\author*[2]{\sur{Wei Lu}}\email{luwei@tju.edu.cn}
%\equalcont{These authors contributed equally to this work.}

\author[3]{\sur{Gang Yu}}%\email{kaihu@tju.edu.cn}
\author[3]{\sur{Bin Fu}} %\email{kaihu@tju.edu.cn}

\affil[1]{\orgdiv{School of Microelectronics}, \orgname{Tianjin University}, \orgaddress{\city{Tianjin}, \postcode{3000072}, \country{China}}}

\affil[2]{\orgdiv{School of Electrical and Information}, \orgname{Tianjin University}, \orgaddress{\city{Tianjin}, \postcode{3000072}, \country{China}}}

\affil[3]{\orgdiv{GY-Lab}, \orgname{Tencent PCG}, \orgaddress{\city{Shanghai}, \postcode{200000}, \country{China}}}
%%==================================%%
%% sample for unstructured abstract %%
%%==================================%%

\abstract{Facial semantic guidance (including facial landmarks, facial heatmaps, and facial parsing maps) and facial generative adversarial networks (GAN) prior have been widely used in blind face restoration (BFR) in recent years. Although existing BFR methods have achieved good performance in ordinary cases, these solutions have limited resilience when applied to face images with serious degradation and pose-varied (e.g., looking right, looking left, laughing, etc.) in real-world scenarios. In this work, we propose a well-designed blind face restoration network with generative facial prior. The proposed network is mainly comprised of an asymmetric codec and a StyleGAN2 prior network. In the asymmetric codec, we adopt a mixed multi-path residual block (MMRB) to gradually extract weak texture features of input images, which can better preserve the original facial features and avoid excessive fantasy. The MMRB can also be plug-and-play in other networks. Furthermore, thanks to the affluent and diverse facial priors of the StyleGAN2 model, we adopt it as the primary generator network in our proposed method and specially design a novel self-supervised training strategy to fit the distribution closer to the target and flexibly restore natural and realistic facial details. Extensive experiments on synthetic and real-world datasets demonstrate that our model performs superior to the prior art for face restoration and face super-resolution tasks.
}


%%\pacs[JEL Classification]{D8, H51}

%%\pacs[MSC Classification]{35A01, 65L10, 65L12, 65L20, 65L70}

\maketitle

\section{Introduction} %\label{sec1}
Face images collected by cameras are often affected by the combination of multiple unknown degradation factors in the wild, such as low resolution, noises, blur\cite{lyu2021boosting}, compression artifacts, etc. Therefore, this may lead to severe loss of color information and blurring of texture details of face images, so the visual sensory quality is lower than before and can not be applied again. Blind face restoration (BFR) is a typically ill-posed inverse problem and aims at reproducing realistic and reasonable high-quality (HQ) face images from unknown degraded inputs. It has been of concern based on the needs of real life (e.g., video/image processing and Augmented Reality\cite{minaee2022modern}). Furthermore, due to the complexity and diversity of facial poses, restoring face images with natural and high-quality results for BFR tasks is still challenging. 

The human face has a highly complex structure and special properties different from other objects. Previous works based on Convolutional Neural Networks (CNN) and GAN use various facial semantic priors (including facial landmarks\cite{chen2018fsrnet,sharma20213Dface_landmark,9591403}, facial parsing maps\cite{richardson2021psp,chen2021psfr-gan}, facial heatmaps\cite{bulat2018super}, and facial component dictionaries\cite{li2020DFD,li2022learning,wang2022restoreformer}) to guide the networks to recover face shape and details. However, when these facial priors are adopted to restore complex degraded images, there is still much space for improvement in the restoration results due to the limited prior information. As a generative facial model with excellent performance, StyleGAN\cite{karras2019stylegan,karras2020stylegan2} is capable of synthesizing face images with rich textures and realistic vision and providing rich and diverse priors, such as facial contours and textures of all areas (including hair, eyes, and mouth). So far, StyleGAN2\cite{karras2020stylegan2} has been widely applied in face restoration tasks, such as PULSE\cite{menon2020pulse}, pSp\cite{richardson2021psp}, PSFR-GAN\cite{chen2021psfr-gan}, GFPGAN\cite{wang2021GFPGAN}, GPEN\cite{yang2021gpen}, GCFSR\cite{he2022gcfsr}, and Panini-Net\cite{wang2022panini}. However, due to the low utilization of texture information for the input image and the lack of reasonable and efficient training strategies for the pre-trained StyleGAN2 model, there is still limited resilience when applied in serious degradation and pose-varied face images.

\begin{figure}[!t] 
	\centering
	\includegraphics[width=3.5in]{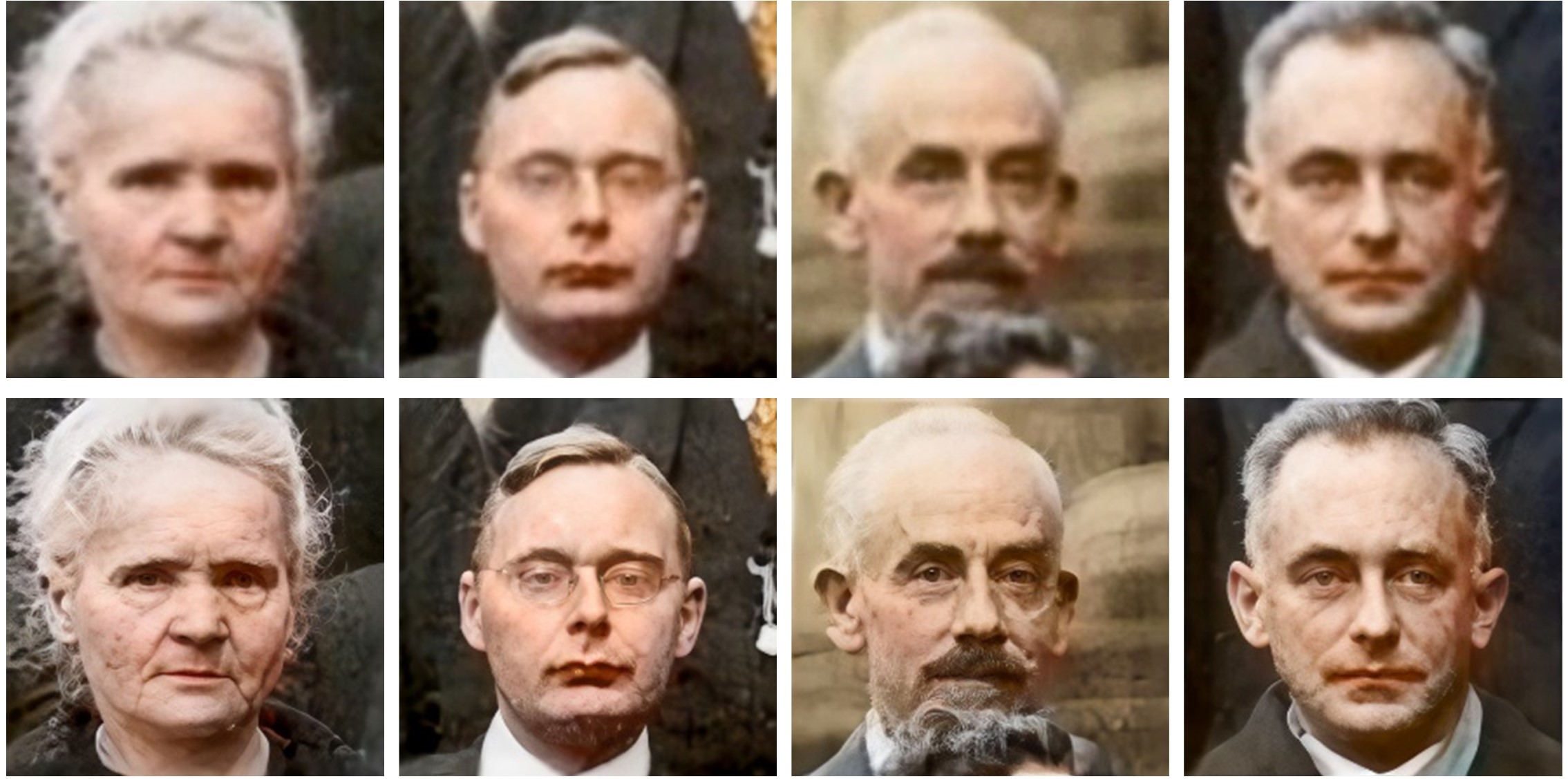}
	\caption{Restoration results for the old photos of the Solvay conference in 1927. Only some face images are displayed. Our method can restore the details of the original images to a large extent and avoid excessive fantasy.  \textbf{Please zoom in for the best view}.}
	\label{fig0:show results}
\end{figure}
To address this challenge, we propose a novel blind face restoration network with the GAN prior, composed of an asymmetric codec and a StyleGAN2 prior network\cite{karras2020stylegan2}. Firstly, since there are few weak texture features in degraded images, we propose a mixed multi-path residual block (MMRB), which mainly adopts a two-branch sparse structure to extract the features of different scales and can achieve spatial interaction and aggregation of shared features through skip connections. In this work, we apply MMRB layers to gradually extract weak texture features of input images from the shallower to the deeper. 

Secondly, in order to better restore natural and high-quality face images, as shown in Fig.\ref{fig0:show results}, we adopt the StyleGAN2 model with generative facial prior as the primary generator network and jointly fine-tune the generator with the codec. Using facial landmark points of training data, the position coordinates of the local areas are obtained and used to crop facial components, such as the eyes and mouth. We then adopt local facial loss to promote the authenticity of the output facial results during training. Finally, we introduce a training strategy of freezing a pre-trained discriminator (FreezeD)\cite{mo2020FreezedD}, which helps our network recover more reasonable and high-fidelity results in complex degraded scenes. In particular, this training strategy helps our model to remain stable during training and speed up the fitting of the generated results to the target distribution. 

Our main contributions are summarized as follows:
\begin{itemize}
\item We propose a well-designed blind face restoration network with generative facial prior, which can promote the quality of face images with complex facial postures and serious degradation. 

\item For the StyleGAN2 prior model, we specially design a novel self-supervised training strategy that freezes a pre-trained discriminator (FreezeD) and jointly fine-tunes the generator with the codec. It helps our model recover high-quality face images more realistically and reasonably and maintain stability during training. 

\item To better extract few and weak texture features in low-quality images, we propose an MMRB layer, which adopts a two-branch sparse structure to extract the features of different scales and can achieve spatial interaction and aggregation of shared features through skip connections. 

\item Our method achieves state-of-the-art results on multiple datasets. It can be observed that our method has good generalization capability and can tackle serious face image degradation in diverse poses and expressions. 
\end{itemize}

\section{Related Work}
\subsection{Face Restoration}
For blind face restoration in the real world, the restoration problem becomes more complex due to the particularity of the face image itself and the influence of many unknown degradation factors. This task has three main research trends\cite{22face_survey, jiang2021faceSR_survey} for this task: basic CNN-based, GAN-based, and Prior-guided methods.

\textit{\textbf{Basic CNN-based methods.}} Chen et al. \cite{chen2018fsrnet} utilized a two-stage network to achieve face super-resolution gradually from coarse to fine based on the facial geometry priors. Xin et al. \cite{xin2019residual} made full use of facial prior information at pixel level and semantic level based on an end-to-end Residual Attribute Attention Network to fulfill high-quality face super-resolution. Xin et al. \cite{xin2020facial} adopted an integrated representation model of facial information to solve the high-scale super-resolution problem in noisy facial images. DPDFN \cite{jiang2020dual} employed a two-branch network to learn the face's local details and global contours without requiring additional face priors respectively. Wang et al. \cite{wan2022old} regarded image restoration as a domain translation problem and proposed two variational autoencoders to transform old and clean photos into two latent spaces and learn latent feature translation to restore high-quality images. Through mapping a low-quality(LQ) image to a codebook space, Zhou et al. \cite{zhou2022towards} exploited a transformer-based code prediction network to improve the quality and fidelity of face restoration. Hu et al. \cite{9591403} adopted a 3D facial prior such as rich hierarchical features, and used a skin perception loss function to promote the performance of face restoration. 

\textit{\textbf{GAN-based methods}}. HiFaceGAN\cite{yang2020hifacegan} proposed a collaborative suppression and replenishment framework to tackle unconstrained face restoration problems. PSFR-GAN\cite{chen2021psfr-gan} proposed a semantic-aware style transfer approach based on the StyleGAN2 network to make use of multi-scale inputs and recover HQ face details progressively. GCFSR\cite{he2022gcfsr} proposed a generative and controllable face super-resolution model without reliance on any additional prior, which can be used to reconstruct faithful images with promising identity information. However, face restoration methods based on CNN and GAN have limited resilience due to lacking generative facial prior.

\textit{\textbf{GAN Prior-guided methods}}. The pre-trained Style-GAN2\cite{karras2020stylegan2} model has been widely used in face restoration and super-resolution tasks due to its rich facial prior information. PULSE\cite{menon2020pulse} adopted a self-supervised training method to iteratively optimize the latent codes of the StyleGAN2 model until the distance between outputs and inputs was below a threshold to restore high-quality facial images. pSp\cite{richardson2021psp} utilized the pre-trained StyleGAN2 model and used a standard feature pyramid as an  encoder network to solve Image-to-Image translation tasks. GFPGAN\cite{wang2021GFPGAN} also used the pre-trained StyleGAN2 model as a generative facial prior and then adopted Channel-split Spatial feature Transform layers to perform spatial modulation on a split of features to improve the quality of face images. GPEN\cite{yang2021gpen} directly embedded GAN prior into the codec structure as the decoder and jointly fine-tuned the GAN prior network with the deep neural network. Panini-Net\cite{wang2022panini} proposed a learnable mask to dynamically fuse the encoder's features with the features generated by GAN blocks in the pre-trained StyleGAN2 model. Although HQ face images can be restored using the GAN prior, their restoration results are not realistic enough for the face images with multiple poses and serious degradation. 

\subsection{Transferring GAN Priors} 
Transferring GAN models is also regarded as a domain adaptation, which mainly adjusts the data distribution of the pre-trained model to a domain suitable for other tasks\cite{zhou2021domain_generalization_survey,karras2020training_generative}. For the pre-trained model, commonly used strategies are to fine-tune or fix the parameters of the pre-trained model\cite{huang2021unsupervised,menon2020pulse}. Most relevant to our work, domain adaptation methods (\cite{tov2021designing}) demonstrate impressive visual quality and semantic interpretability built upon the StyleGAN in the target domain. StyleGAN-ADA\cite{karras2020training_generative} proposed an adaptive discriminator augmentation method to train the StyleGAN network on limited data samples. Mo et al.\cite{mo2020FreezedD} froze lower layers of the discriminator to achieve domain adaptation and demonstrated the effectiveness of this simple baseline using various architectures and datasets. Huang et al.\cite{huang2021unsupervised} adopted an unsupervised Image-to-Image translation method to generate multi-domain face images through the pre-trained StyleGAN2 model.

\subsection{Feature Extraction Block}
So far, various feature extraction blocks have been proposed to improve the learning ability of convolution layers in a network. Multiple versions of inception blocks\cite{szegedy2015going}\cite{szegedy2017inception} adopted parallel multi-scale convolution layer in the network to replace the dense structure, fused all features, and used the bottleneck layer for dimension reduction. Kim et al.\cite{he2016deep} added a short connection to the previous convolution layer and put forward the idea of residual learning, which can effectively alleviate the disappearance of gradients during training and reduce the number of parameters. After that, Zhang et al.\cite{zhang2018residual} mainly increased the width of convolution by using the information compensation of other feature channels, which can reduce the vanishing gradient and realize the efficient utilization of feature information. Motivated by the inception blocks, Li et al.\cite{li2018multi-scale} proposed a multi-scale residual block to exploit the features of images at different scales. Inspired by the above methods, we propose a mixed multi-path residual block (MMRB) to extract image features of different paths and realize mixed interaction among these to provide better modeling capabilities for face restoration. 
\begin{figure*}[!t] 
	\centering
	\includegraphics[width=1.0\textwidth]{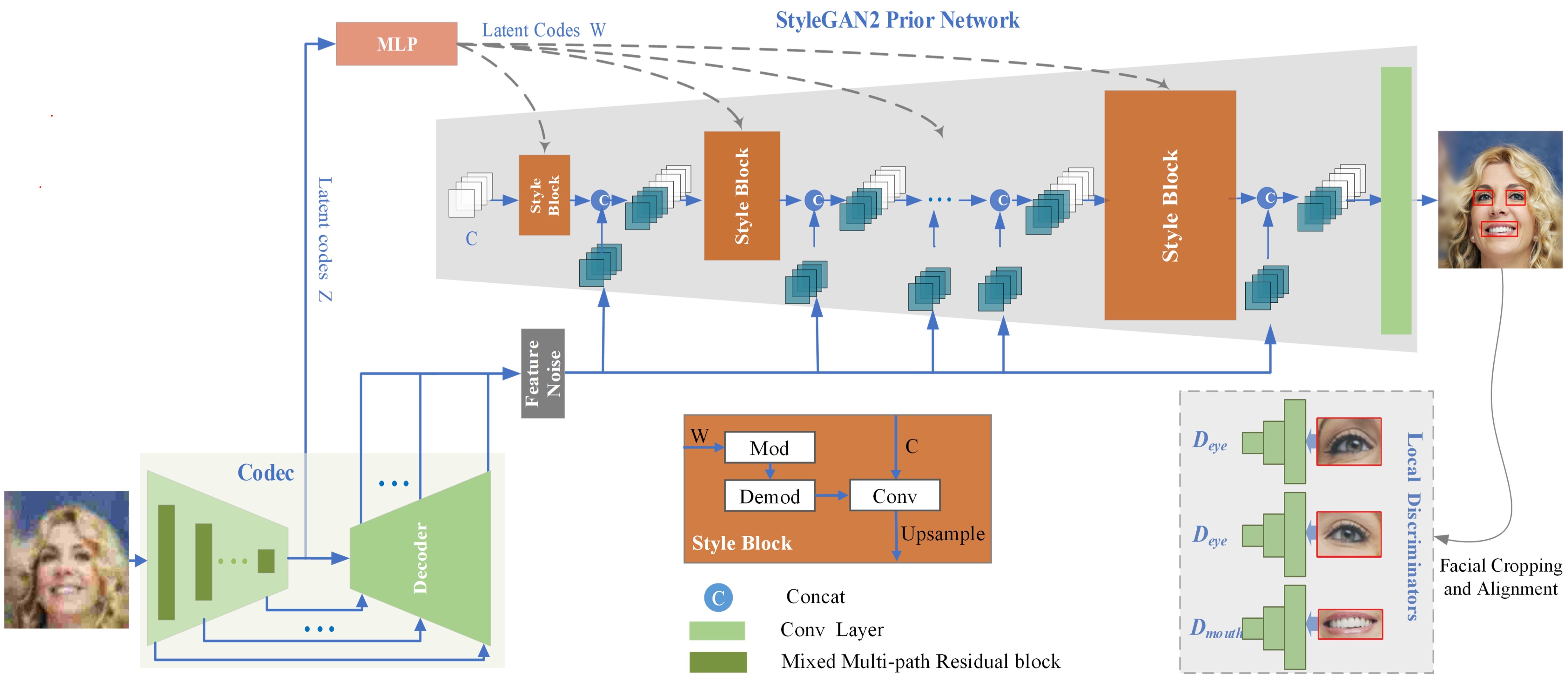} %[width=2.5in]
	\caption{The framework of our proposed method. It is mainly composed of an asymmetric codec and a StyleGAN2 prior network. There is only one MMRB layer for each convolution scale and a total of 6, except for the encoder's first and last convolution scales. Furthermore, the inputs of the GAN prior network include latent codes $\boldsymbol{W}$, a learned 4 × 4 × 512 constant tensor $\boldsymbol{C}$, and noise branches. This prior network can then apply style blocks to restore high-quality face images from coarse to fine gradually.}
	\label{fig1:network}
\end{figure*}

\section{Our Proposed Methods}
\subsection{Global Architecture}
In this section, we will describe the well-designed network framework in Fig.\ref{fig1:network}, which can be used to solve a serious ill-posed inverse problem in the wild. This framework integrates the excellent strategies of the GFPGAN\cite{wang2021GFPGAN} and GPEN\cite{yang2021gpen} methods, mainly focusing on embedding multi-scale features of input images into the StyleGAN2 generator network. It is also inspired by StyleGAN2\cite{karras2020stylegan2} and consists of an asymmetric codec and a StyleGAN2 prior network. Firstly, We input a low-quality image $\boldsymbol{x}$ into the codec, in which the encoder has one more MMRB layer in each scale than the decoder. That is because the encoder needs to provide more natural and reliable face latent features for the generator. The MMRB layer proposed by us can better extract the degraded images' weak and few texture features. The encoder then maps the input image $\boldsymbol{x}$ to the closest latent codes $\boldsymbol{Z}$ in StyleGAN2. It can be seen from the StyleGAN\cite{karras2019stylegan} and StyleGAN2\cite{karras2020stylegan2} that images can be generated only by latent codes without noise branches, but the generated results are quite different from the input images and lack rich and real texture information. We have also verified and found that the reconstructed feature maps have richer and more reasonable facial textures than the direct output of the encoder during training. Aiming to explicitly remove existing degraded factors and extract ‘clean’ features, we use L1 restoration loss similar to\cite{wang2021GFPGAN} for each resolution scale of the reconstructed feature maps in the decoder and directly output the feature maps at each scale to the noise branches in the StyleGAN2. Thus, the process of codec is formulated as $:$
\begin{align}
	\boldsymbol{\hat{x}}, \boldsymbol{Z} = Codec \left( \boldsymbol{x} \right).
\end{align}

Secondly, latent codes $\boldsymbol{W}$, a learned constant feature tensor $\boldsymbol{C}$ and noise features $\boldsymbol{\hat{x}}$ are input into the pre-trained StyleGAN2, where $\boldsymbol{W}$ are latent codes decoupled by the latent codes $\boldsymbol{Z}$ through a mapping network of using an 8-layer multi-layer perception (MLP). Decoupling feature space is formulated as$:$
\begin{align}
	\boldsymbol{W} = MLP \left( \boldsymbol{Z} \right).
\end{align}
The $\boldsymbol{W}$ ($\boldsymbol{w} \in \boldsymbol{W}$) is then broadcast to each style block, in which the Mod and Demod indicate the modulation and demodulation operation of latent codes $\boldsymbol{W}$, respectively. In particular, the input noises $\boldsymbol{\hat{x}}$ and the output $\boldsymbol{y}$ of the style block after feature modulation are fused by concatenating rather than directly adding to convolutions in the StyleGAN2 model, which can take advantage of the features introduced by noise branches to restore the texture of local areas flexibly. 
\begin{align}
    \boldsymbol{y}_{i+1} = Style \left( \boldsymbol{w}, Concat \left[ \boldsymbol{\hat{x}}_{i}, \boldsymbol{y}_{i} \right] \right).
\end{align}
The fused feature maps $\boldsymbol{y}$ are fed to the next style block. On the basis of fine-tuned training skills, we can apply style blocks to restore high-quality face images from coarse to fine gradually.

\subsection{Mixed Multi-path Residual block}
\begin{figure}[!t]
    \centering
    \includegraphics[width=3.5in]{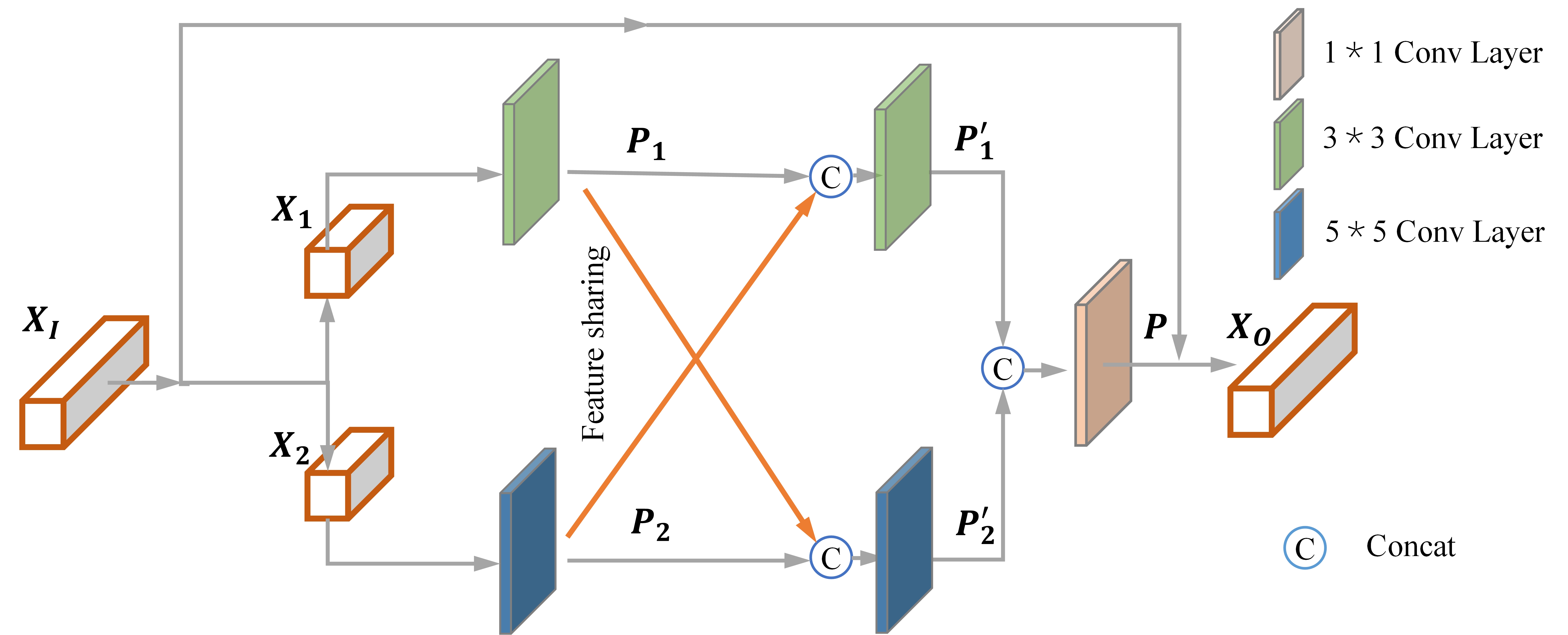}
    \caption{Overview of the mixed multi-path residual block (MMRB).}
    \label{fig2:MMRB}
\end{figure}
To better preserve the original facial features and avoid excessive fantasy, we propose a mixed multi-path residual block (MMRB) to extract slight and weak high-frequency textures in the degraded image. Here we will clearly describe its structure in Fig.\ref{fig2:MMRB}. We design a two-branch and interactive feature extraction network, unlike previous works\cite{szegedy2015going,szegedy2017inception,zhang2018residual,li2018multi-scale}. We first equally split the input multi-channel feature maps $\boldsymbol{X}_{I}$ rather than reduce dimension using a $ 1 \times 1 $ bottleneck layer to reduce introduced parameters. 
\begin{align}
	\boldsymbol{X}_{1}, \boldsymbol{X}_{2} = Split \left( \boldsymbol{X}_{I} \right).
\end{align}
Where $Split \left(  \cdot \right)$ represents the separation operation of the feature maps in the channel dimension. $\boldsymbol{X}_{I} \in {\mathbb{R}}^{H \times W \times C}$.  $\boldsymbol{X}_{1},\boldsymbol{X}_{2} \in {\mathbb{R}}^{H \times W \times C {/} 2}$.
We then adopt the two-branch sparse structure to extract the features of different scales. Skip connections are applied to share extracted features and achieve spatial interaction and aggregation of different features. The detailed definitions are as follows:
\begin{align}
	\boldsymbol{P}_{1} = \boldsymbol{F}_{1} \left( \boldsymbol{X}_1, \boldsymbol{W}_1 \right) = \sigma \left( \boldsymbol{W}_1 \left( \boldsymbol{X}_1 \right) \right),
\end{align}
\begin{align}
	\boldsymbol{P}_{2} = \boldsymbol{F}_{2} \left( \boldsymbol{X}_2, \boldsymbol{W}_2 \right) = \sigma \left( \boldsymbol{W}_2 \left( \boldsymbol{X}_2 \right) \right),
\end{align}
%\begin{align}
%    C_{1} = Concat \left[ P_1, P_2 \right],
%\end{align}
\begin{align}
	\boldsymbol{P}^{'}_{1} = \boldsymbol{F}_{1} \left( \boldsymbol{C}_1, \boldsymbol{W}_1 \right) = \sigma \left( \boldsymbol{W}_1 \left( Concat \left[ \boldsymbol{P}_1, \boldsymbol{P}_2 \right] \right) \right),
\end{align}
\begin{align}
	\boldsymbol{P}^{'}_{2} = \boldsymbol{F}_{2} \left( \boldsymbol{C}_1, \boldsymbol{W}_2 \right) = \sigma \left( \boldsymbol{W}_2 \left( Concat \left[ \boldsymbol{P}_1, \boldsymbol{P}_2 \right] \right) \right),
\end{align}
%\begin{align}
%    C_{2} = Concat \left[ P^{'}_1, P^{'}_2 \right],
%\end{align}
\begin{align}
	\boldsymbol{P} = \boldsymbol{F}_{3} \left( \boldsymbol{C}_2, \boldsymbol{W}_3 \right) = \sigma \left( \boldsymbol{W}_3 \left( Concat \left[ \boldsymbol{P}^{'}_1, \boldsymbol{P}^{'}_2 \right] \right) \right).
\end{align}
Where $\boldsymbol{F} \left( \cdot , \cdot \right)$ represents convolution mapping functions, $\sigma \left( \cdot \right)$ stands for the PReLU nonlinear activation function\cite{he2015delving} , $ \boldsymbol{W}_{1}, \boldsymbol{W}_{2}, \boldsymbol{W}_{3}$ indicate that the convolution kernel size used in the convolution layer are 3, 5, and 1, respectively and $ Concat \left[\cdot , \cdot \right]$ denotes the concatenation operation. Finally, the output feature parameters are significantly reduced through the dimension reduction of  $ 1 \times 1 $ bottleneck layer. 

In order to make it more efficient and practical, we adopt residual learning for each MMRB. Formally, we describe the MMRB layer as:
\begin{align}
    \boldsymbol{X}_O = \boldsymbol{X}_I + \boldsymbol{P}.
\end{align}
Where $\boldsymbol{X}_O $ represents the MMRB's output. $\boldsymbol{P} \in \mathbb{R}^{H \times W \times C}$ and is the fused output of multi-path feature maps. The MMRB layer is applied in the encoder level by level from the shallow layer to the deep layer to improve our network's performance. What is more, it introduces fewer parameters and can be plug-and-play to enhance the ability of feature extraction in other networks. 

\subsection{Fine-tuning and FreezeD strategies}
Thanks to the powerful generation ability of the StyleGAN2\cite{karras2020stylegan2} prior model, we use the StyleGAN2 model as a main prior network. We adopt different learning rate strategies for the codec and the pre-trained StyleGAN2 network instead of fixing its weights like other methods to achieve joint fine-tuning training. The main idea of fine-tuning is to adopt a lower learning rate, which allows the weight parameters of the StyleGAN2 prior network to be slightly updated during each iteration process to fit the optimal results gradually. While the source and target domains are the same in our work, fine-tuning the StyleGAN2 network can better fit the complex degraded multi-pose face data into the target distribution. To make the distribution of generating results more consistent with the distribution of targets, we introduce a FreezeD training skill\cite{mo2020FreezedD} to fine-tune the StyleGAN2 network by freezing the higher-resolution layers of the discriminator during training. We find that simply freezing the lower layers of the discriminator and only fine-tuning the upper layers performs surprisingly well, and the generator is more stable during training.  

\subsection{Model Objectives}
To jointly fine-tune our restoration model, we use the following loss functions: the adversarial loss, the reconstruction loss, and the face-preserving loss. The adversarial loss is composed of a global adversarial loss and multiple local adversarial losses, in which the global adversarial loss function is defined as 
\begin{equation}
	L_{adv-g} = \min_{G} \max_{D} \mathbb{E}_{(X)} \log \left(1 + \exp \left(- D \left(G \left( \boldsymbol{X}^{'} \right)\right)\right)\right).
\end{equation}
Where $\boldsymbol{X}$ and $\boldsymbol{X}^{'}$ denote the ground-truth image and the generated one, and G is the pre-trained StyleGAN2 for fine-tuning. D is the pre-trained discriminator model from\cite{yang2021gpen} and adopts the FreezeD strategy during training. 

We believe the StyleGAN2\cite{karras2019stylegan} model incorporates geometric face priors for BFR tasks. Therefore, we introduce local facial losses to focus generated results on local patch distributions, such as eyes and mouth. By using these local losses, we can effectively distinguish the high-frequency features of local regions to generate more natural and vivid local content. In particular, it can improve our model to solve the unreal problem of multi-pose face restoration in the wild. 

We employ local discriminators for the left eye, right eye, and mouth. Here local discriminator network is similar to that of the global discriminator. Due to the overall small size of the cropped image patch for each facial component, we only utilize two down-sampling operations in the local discriminator and finally output a single-channel feature map to calculate the discriminative loss. Furthermore, the facial components $X^{'}_{comp}$ need to be cropped and aligned based on the 68 key points solved by the facial key points detection model\cite{jin2021facial_landmark} and Mask R-CNN\cite{he2017mask}. We build upon the same discriminator network for each region and employ different local losses for adversarial training. The local facial losses are defined as:
\begin{equation}
	L_{adv-l} = \sum^{N=3}_{i=1} \left[ \min_{G} \max_{D}\mathbb{E}_{(X^{'}_{comp(i)})} \log \left(1 - D_{comp(i)} \left( \boldsymbol{X}^{'}_{comp(i)} \right)\right)\right],
\end{equation}
\begin{equation}
	L_{adv} = \lambda_{g} L_{adv-g} + \lambda_{l} L_{adv-l}. 
\end{equation}
Where the cross-entropy loss function is used, and $D_{comp}$ is the local discriminator for each region. $ \lambda_{g}$ and $ \lambda_{l}$ represent the loss weights of global adversarial loss and local adversarial losses, respectively. We differentiate the weights of the discriminators and set the weights of local discriminators to 6 times of the global ones to force the generator to focus more on local regions rather than the whole image during training. Besides, we adopt the L1-norm loss and feature matching loss as content losses $L_{C}$:
\begin{equation}
	L_{C} = \lambda_{L1} \left\lVert \boldsymbol{X} - \boldsymbol{X}^{'} \right\rVert_1 + \lambda_{FM} \left[ \min_{G} \mathbb{E} \left( \sum^N_{i=1}  \left\Vert \psi_{i} \left(  \boldsymbol{X} \right) - \psi_{i} \left( \boldsymbol{X}^{'} \right) \right\rVert_2 \right) \right],
\end{equation}
$\lambda_{L1}$ and $ \lambda_{FM}$ represent the loss weights of the L1-norm loss and feature matching loss\cite{wang2018high}. Inspired by\cite{zhao2016loss_functions}, we give $ \lambda_{FM}$ a very small value of the weight parameter to avoid generating smooth results and checkerboard artifacts. $\psi_{i}$ indicates the $i$-th convolution layer of the pre-trained VGG network\cite{simonyan2014very}. $N$ is the total number of intermediate layers used for feature extraction. To enforce the restored face to have a small distance with the ground truth in the deep feature space, we introduce a face-preserving loss\cite{huang2017beyond}:
\begin{equation}
	L_{FP} =  \lambda_{FP}  \left\Vert \phi \left( \boldsymbol{X} \right) - \phi \left( \boldsymbol{X}^{'} \right) \right\rVert_1.
\end{equation}
where $\lambda_{FP}$ denotes the loss weight. $\phi$ represents the face feature extractor that adopts the pre-trained ArcFace\cite{jiankang2019retinaface} model for face recognition.

The overall loss optimization function used by our model is defined as 
\begin{equation}
	L_{total} = L_{adv} + L_{C} +  L_{FP}.
\end{equation}
All the above hyper-parameters are set as follows: $\lambda_g = 0.5$, $\lambda_l = 3$, $\lambda_{L1} = 8$, $\lambda_{FM} = 0.02$, $\lambda_{FP} = 10$.

\section{Experiments}
\subsection{Datasets}
\textit{\textbf{Training Datasets}}. We train on the 70k high-quality face images from the FFHQ dataset, which is synthesized by\cite{karras2020stylegan2}. We resize the resolution of all images to 512$\times$512 during training. To make the training data more in line with the degradation of the real scenes, we follow the practice in\cite{wang2021real-esrgan,wang2021GFPGAN,yang2021gpen} and adopt the following degradation model with all possible degradation factors in the wild to synthesize LQ training data: 
\begin{equation}
	\boldsymbol{x} = \mathbb{D} (\boldsymbol{y}) = \left[ \left(\boldsymbol{y} \otimes \boldsymbol{k} \right) \downarrow_{\boldsymbol{s}} + \boldsymbol{n}_\delta \right]_{JPEG_q},
\end{equation}
$\mathbb{D}$ denotes the degradation process for HQ images. The HQ face image $\boldsymbol{y}$ is first convolved $\otimes$ with blur kernel $\boldsymbol{k}$. Then, a down-sampling operation with scale factor $\boldsymbol{s}$ is performed. Meanwhile, we continue to superimpose adding noise $\boldsymbol{n}$ on the high-quality image. Finally, the LQ image $\boldsymbol{x}$ is obtained by $\emph {JPEG}$ compression. To simulate the severe degraded scenes, we randomly sample $\boldsymbol{k}$, $\boldsymbol{s}$, $\boldsymbol{\delta}$ and $\boldsymbol{q}$ from $41$, $[0.4 : 8]$, $[0 : 25]$, $[5 : 50]$ for each training pair in our experiments, respectively. In particular, we use Gaussian blur and motion blur to deal with the possible blur of the image in the natural scenes.

\textit{\textbf{ Testing Datasets}}. To better verify the performance of our model in complex degraded scenes, we use various types of low-quality datasets for testing, and they do not overlap with the training data. We make a brief description of these datasets.

\textit{1) CelebA Data} is a synthetic dataset with 30000 high-quality face images\cite{liu2015deep}. We select 2556 face images with complex facial postures and degrade the chosen images using the above degradation model and parameters.

\textit{2) LFW Data} comes from\cite{huang2008LFW} and is composed of more than 13k face pictures of famous people all over the world with different orientations, expressions, and lighting environments. They are really low-quality images with only a single face. The image size is 250 pixels $\times$ 250 pixels. We select 1610 images with relatively low quality for testing.

\textit{3) FDDB Data} comes from\cite{fddbData} and contains 5171 human faces in 2845 images taken from different natural scenes. The number of faces in a single image is not unique. Moreover, the width and height of the images are not equal, and the difference is noticeable. We select 1629 images with relatively low quality for testing.

\textit{4) WebFace Data} is collected face images from the Internet by\cite{yi2014Webface}, including tens of thousands of images with a size of 250 pixels $\times$ 250 pixels. They are low-quality images with only a single face and many old black-and-white photos. We select 804 images with relatively low quality for testing.

\subsection{Implementation Details and Evaluation Metrics}
We adopt the FreezeD strategy\cite{mo2020FreezedD} to the global discriminator from\cite{yang2021gpen}, freeze the parameters of the first five convolution layers, and fine-tune the remaining deep features. We then jointly fine-tune the Stylegan2 model retrained by\cite{yang2021gpen} with the codec. This is because we are consistent with the design of the generator model between the GPEN method\cite{yang2021gpen}, which has pre-trained generator and discriminator models similar to the initial StyleGAN2 before doing the BFR task. The input image dimension mapped to the nearest latent codes is consistent with the pre-trained Stylegan2 model in our work. At the same time, the resolution and dimension for the output feature maps of every level in the reconstruction process are consistent with the noise branch of the Stylegan2 model. For each MMRB layer, we only adopt one for each convolution scale and a total of 6, except for the encoder's first and last convolution scales. 
\begin{figure*}[!t] 
	\centering
	\includegraphics[width=1.0\textwidth]{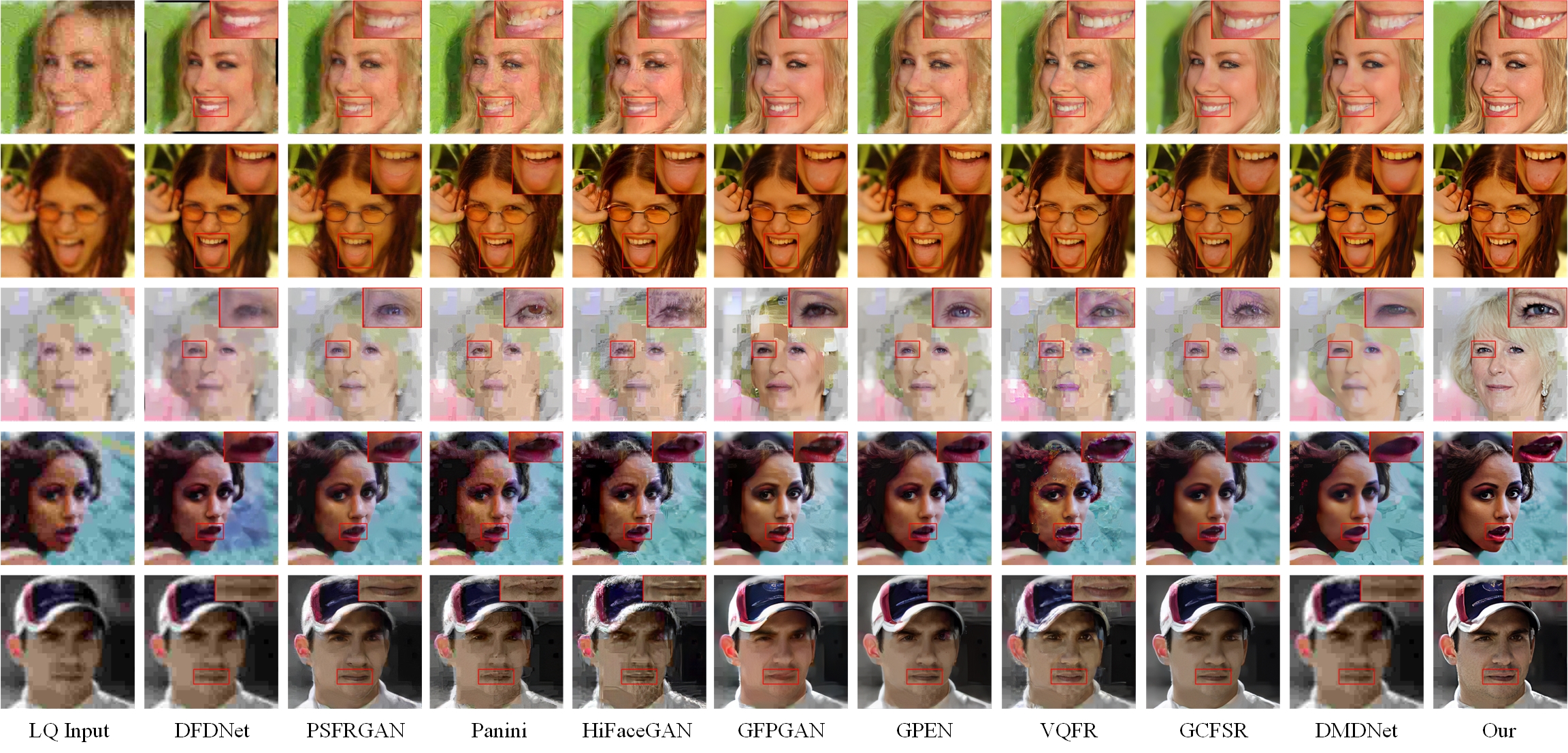}
	\caption{Qualitative comparisons with several state-of-the-art face restoration methods on the CelebA Data. To intuitively feel the performance difference of each method, we enlarge and display local areas. \textbf{Zoom in for best view}.}
	\label{fig: BFR celeba results}
\end{figure*}

During the experiment, we train our model with the Adam optimizer\cite{kingma2014adam} and perform a total of 700k iterations with a batch size of 4. The learning rate (LR) varies for different parts, including the codec, the pre-trained Stylegan2 model, noise branches, the local discriminators, and the global discriminator. They are set to $2\times10^{-3}$, $2\times10^{-4}$, $2\times10^{-3}$,  $2\times10^{-3}$, and $2\times10^{-5}$, respectively. The local discriminators include three facial components: left eye, right eye, and mouth. They use the same learning rate and discriminator model. Furthermore, a piece-wise attenuation strategy is adopted. We implement our model with the PyTorch framework and train it using two NVIDIA V100 GPUs.

For the evaluation, we employ two widely-used non-reference perceptual metrics: NIQE\cite{mittal2012NIQE} and IFQA\cite{jo2023ifqa} (Interpretable Face Quality Assessment). Their evaluation of facial image quality is exactly complementary. Moreover, we also adopt pixel-wise metrics (PSNR and SSIM) and the reference perceptual metrics (LPIPS\cite{zhang2018LPIPS} and FID\cite{2017FID}) for the CelebA Data with Ground Truth (GT). 

\subsection{Experiments on Synthetic Images}
To better show the practicability and generality of our model, we verify the performance of face restoration and face super-resolution tasks through synthetic CelebA Data. 

\begin{table}%[]
	%\vspace{-0.3cm}
	%\small
	\centering
	\caption{Quantitative comparisons of various BFR methods on \textbf{CelebA Data}. Bold \textcolor{red}{\bf RED} indicates the best performance, and bold \textcolor{blue}{\bf BLUE} indicates the second. Reference evaluation metrics(e.g., PSNR, SSIM, LPIPS, and FID) are adopted, and the non-reference perceptual metrics (e.g., NIQE, IFQA) are adopted.}
	\label{tab:celeba_BFR}
	\scalebox{1.0}{
		%\hspace{-0.5cm}
    	\begin{tabular}{c|cc|cc|cc}
    		\hline
    		Methods     & PSNR$\uparrow$  & SSIM$\uparrow$  & LPIPS$\downarrow$  & FID$\downarrow$ &NIQE $\downarrow$ & IFQA$\uparrow$     \\ \hline
    		HiFaceGAN~\cite{yang2020hifacegan}  & 24.506  & 0.612 & 0.196  & 31.545 & 4.427  & 0.279\\
    		DFDNet~\cite{li2020DFD}    & 22.470 & 0.663 & 0.232   & 54.358 & 5.916  & 0.166  \\ %\blueud{}
    		VQFR~\cite{gu2022vqfr}    & 24.907  & 0.675 & 0.175   & 21.312 & 4.580 & 0.313   \\ %\blueud{}
    		PSFRGAN~\cite{chen2021psfr-gan}    & 24.884 & 0.663  & 0.192  & 34.695 & 4.765 & 0.151    \\ %\hline
    		Panini~\cite{wang2022panini}   & 24.679    &  0.611  &  0.194   &  44.696   & \textcolor{red}{\bf 3.837} & 0.227 \\
                DMDNet~\cite{li2022learning} & 25.691 & 0.696 & 0.171 &24.871 &5.341 &0.329\\
                GCFSR~\cite{he2022gcfsr} & \textcolor{blue}{25.925} & \textcolor{blue}{0.700} &\textcolor{blue}{0.163} &\textcolor{blue}{16.859} &4.963 &\textcolor{blue}{\bf0.367}  \\
    		GPEN~\cite{yang2021gpen}    & 25.190     &  0.680   & 0.169   &  21.852    &  4.712 & 0.242    \\
    		GFPGAN~\cite{wang2021GFPGAN}   & 24.895  & 0.688   &  0.172     &  21.160  &  4.746  & 0.316 \\\hline
    		\textbf{Our}& \textcolor{red}{\bf 26.034}   & \textcolor{red}{\bf 0.702}    & \textcolor{red}{\bf 0.162}  & \textcolor{red}{\bf 16.080}   & \textcolor{blue}{\bf 4.285} & \textcolor{red}{\bf 0.370}  \\ \hline
    		GT         & $\infty$  & 1  & 0    & 2.478    &  4.188  & 0.430 \\ \hline
	\end{tabular}}
	%\vspace{-0.3cm}
\end{table}
\textit{\textbf{Face Restoration.}} To verify the superiority of the method mentioned in this paper, we make qualitative and quantitative comparisons with several of the latest blind face restoration methods, such as DFDNet\cite{li2020DFD}, VQFR\cite{gu2022vqfr}, PSFRGAN\cite{chen2021psfr-gan}, Panini\cite{wang2022panini}, HiFaceGAN\cite{yang2020hifacegan}, GFPGAN\cite{wang2021GFPGAN}, GPEN\cite{yang2021gpen}, DMDNet~\cite{li2022learning}, and GCFSR~\cite{he2022gcfsr}. We also introduce metric values corresponding to the GT to understand better the difference between the test results of different methods and the GT in quantitative analyses. 

\begin{figure*}[!t] 
	\centering
	\includegraphics[width=1.0\textwidth]{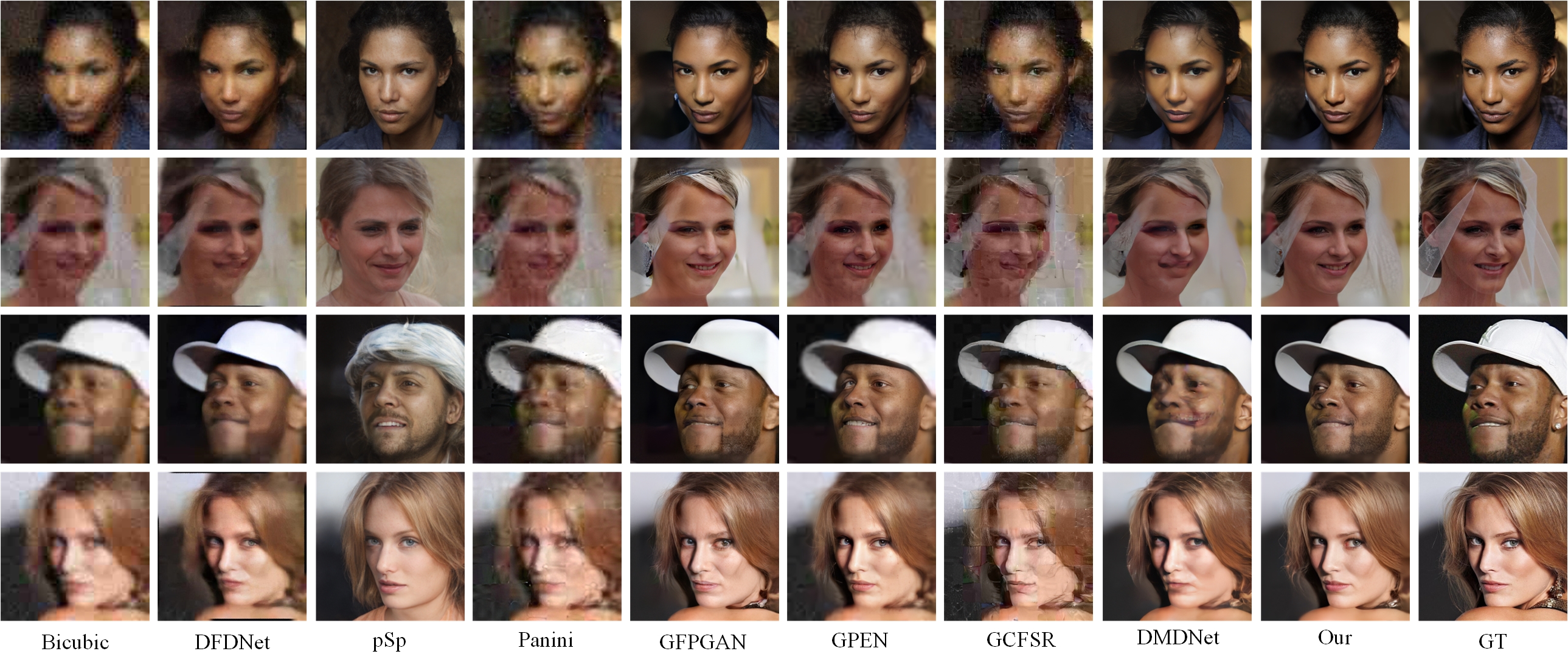} 
	\caption{Qualitative comparisons of 4x face super-resolution by different methods. We do not apply local magnification displays for the qualitative comparison results, which intends to view the differences between them in the form of the whole picture. \textbf{Zoom in for best view}.}
	\label{fig:FSR CelebA results}
\end{figure*}
The quantitative and qualitative results of different methods are shown in Tab.\ref{tab:celeba_BFR} and Fig.\ref{fig: BFR celeba results}, respectively. Our method is optimal in all indicators and performs best according to the quantitative results, and can restore the complex degraded multi-pose face image more flexibly. Furthermore, the IFQA metric\cite{jo2023ifqa} is mainly adopted to evaluate the quality of facial primary regions in face restoration. And the higher the IFQA value, the more reasonable the restored facial structure and the more realistic the details are. Tab.\ref{tab:celeba_BFR} shows that the IFQA value obtained by our method is the highest, closer to GT. The visual results in Fig.\ref{fig: BFR celeba results} show that our restoration images are more realistic and natural, especially in the facial regions(e.g., eyes, mouth). The restored face quality is almost consistent with human visual perception. In particular, the difference can be seen more clearly and intuitively by zooming in on the local details. 

\textit{\textbf{Face Super-resolution.}} We use the first 1500 images of CelebA Data and perform 4x downsampling of the input images to verify the super-resolution performance of different methods in the wild. Furthermore, we also make qualitative and quantitative comparisons with several state-of-the-art face super-resolution methods, including SuperFAN\cite{bulat2018super}, pSp\cite{richardson2021psp}, Panini\cite{wang2022panini}, HiFaceGAN\cite{yang2020hifacegan}, DFDNet\cite{2017FID}, GFPGAN\cite{wang2021GFPGAN}, GPEN\cite{yang2021gpen}, DMDNet~\cite{li2022learning}, and GCFSR~\cite{he2022gcfsr}. Our method needs to upsample tested images to the original scale and then input them into the network for testing.

\begin{table}%[]
	%\vspace{-0.5cm}
	%\small
	\centering
	\caption{Quantitative comparisons of various FSR methods on \textbf{CelebA Data}. We adopt the Bicubic interpolation for 4x down-sampling of the input data to verify the super-resolution performance of different methods. Bold \textcolor{red}{\bf RED} indicates the best performance, and bold \textcolor{blue}{\bf BLUE} indicates the second one.}
	\label{tab:celeba_FSR}
	\scalebox{1.0}{
		%\hspace{-0.5cm}
    	\begin{tabular}{c|cc|ccc}
    		\hline
    		Methods     & PSNR$\uparrow$  & SSIM$\uparrow$  & LPIPS$\downarrow$  & FID$\downarrow$ &NIQE $\downarrow$      \\ \hline
    		Bicubic    & 23.757 & 0.641  & 0.297  & 148.468 & 10.460    \\ %\hline
    		SuperFAN\cite{bulat2018super}    & 23.193 & 0.617  & 0.290  & 152.188 & 7.857    \\ %\hline
    		pSp\cite{richardson2021psp}    & 18.493 & 0.580  & 0.242  & 67.596 & 5.647   \\ %\hline
    		HiFaceGAN\cite{yang2020hifacegan}  & 19.946  & 0.470 & 0.326  & 230.756 & 10.480  \\
    		DFDNet\cite{2017FID}     & 21.772 & 0.638 & 0.260   & 93.107 & 6.416    \\ %\blueud{}
    		Panini\cite{wang2022panini}   & 23.417    &  0.620  &  0.278   &  165.834   & 6.898  \\
    		GPEN\cite{yang2021gpen}   & 24.248     & 0.658  &  0.217   &  50.348    &  5.886     \\
    		GFPGAN\cite{wang2021GFPGAN}   & 23.780   & 0.656  & \textcolor{blue}{0.190}  &\textcolor{blue}{32.894} & 4.955  \\
                DMDNet~\cite{li2022learning} & \textcolor{blue}{24.319} & \textcolor{blue}{0.660} & 0.202 &41.952 &5.869 \\
                GCFSR~\cite{he2022gcfsr} & 23.321  & 0.572  & 0.274  & 118.506 & \textcolor{red}{4.051} \\ \hline
    		\textbf{Our} & \textcolor{red}{\bf 24.797} & \textcolor{red}{\bf 0.674}  & \textcolor{red}{\bf 0.189} &  \textcolor{red}{\bf 27.403} & \textcolor{blue}{\bf 4.449} \\ \hline
    		GT         & $\infty$  & 1  & 0    & 2.413    &  4.036   \\ \hline
	\end{tabular}}
	%\vspace{-0.3cm}
\end{table}
The quantitative and qualitative results of different methods are shown in Tab.\ref{tab:celeba_FSR} and Fig.\ref{fig:FSR CelebA results}, respectively. As can be seen from the quantitative results, our method can perform super-resolution on LQ images, and the output results are significantly better than other face super-resolution methods in all indicators. In particular, our results can be closer to the GT using the MMRB layer, both in qualitative and quantitative results. We do not adopt local magnification displays for the qualitative comparison results, which intends to view the differences between them in the form of the whole picture. By visual comparison results, our results have fewer or less noticeable artifacts and texture blur, whether in facial texture, hair, or background. Although the pSp method\cite{richardson2021psp} also shows promising results in the visualization effect, it is quite different from the GT due to its design idea.

\subsection{Experiments on Images in the Wild}
\begin{table}%[]
	%\vspace{-0.3cm}
	%\small
	\centering
    \caption{Quantitative comparisons of various BFR methods in the wild, such as \textbf{FDDB Data}, \textbf{LFW Data}, \textbf{WebFace Data}. Only the non-reference perceptual metrics (e.g., NIQE, IFQA) are adopted. “\textbf{-}” means that the result is unavailable.}
    \label{tab:BFR_wild}
	\scalebox{1.0}{
	%	\hspace{-0.5cm}
    	\begin{tabular}{c|cc|cc|cc}
    		\hline
    		Dataset       & \multicolumn{2}{c|}{\textbf{FDDB Data}} & \multicolumn{2}{c|}{\textbf{LFW Data}}   & \multicolumn{2}{c}{\textbf{WebFace Data}} \\ 
    		Methods       & IFQA$\uparrow$  & NIQE$\downarrow$  & IFQA$\uparrow$  & NIQE $\downarrow$ & IFQA$\uparrow$ & NIQE $\downarrow$   \\ \hline
    		Input         & 0.265 & 4.262    & 0.298 & 5.454     & 0.251 & 6.197   \\
    		HiFaceGAN\cite{yang2020hifacegan}    &  0.415 & \textcolor{blue}{\bf 3.956}     &  0.377 & 4.548  &  0.367 & 4.969 \\
    		DFDNet\cite{li2020DFD}  &  0.381 & 4.153    &  0.306 & 4.990  &  0.294 & 5.449  \\
    		PSFRGAN\cite{chen2021psfr-gan}     &  0.365 & 4.334  & 0.307 & 5.461  &  0.301 & 5.830 \\
    		Panini\cite{wang2022panini}        &  $-$ & $-$   &  0.367 & 4.694   & 0.376 &  4.875 \\
    		VQFR\cite{gu2022vqfr}         &  0.408  & 4.049   & 0.402  & \textcolor{red}{\bf 4.449}    & 0.404 & \textcolor{red}{\bf 4.390}  \\
    		GPEN\cite{yang2021gpen}     & 0.434  & 4.110  & 0.352  & 5.418   & 0.382  & 5.793  \\
    		GFPGAN\cite{wang2021GFPGAN}    & 0.427 & 4.009   & 0.346  & 4.982    & 0.366 & 5.313  \\
                DMDNet~\cite{li2022learning}    & \textcolor{red}{\bf 0.460} & 4.476   & \textcolor{blue}{\bf 0.415} & 4.733    & 0.418 & 4.898  \\
                GCFSR~\cite{he2022gcfsr}    &  0.428 & 4.979   & 0.404 & 5.136    & \textcolor{blue}{\bf 0.420} & 5.071  \\ \hline
    		\textbf{Ours} & \textcolor{blue}{\bf 0.436} & \textcolor{red}{\bf 3.578} & \textcolor{red}{\bf 0.446} & \textcolor{blue}{\bf 4.548}  & \textcolor{red}{\bf 0.427}  & \textcolor{blue}{\bf 4.605} \\ \hline
	\end{tabular}}
	%\vspace{-0.5cm}
\end{table}

\begin{figure*}[!t] 
	\centering
	\includegraphics[width=1.0\textwidth]{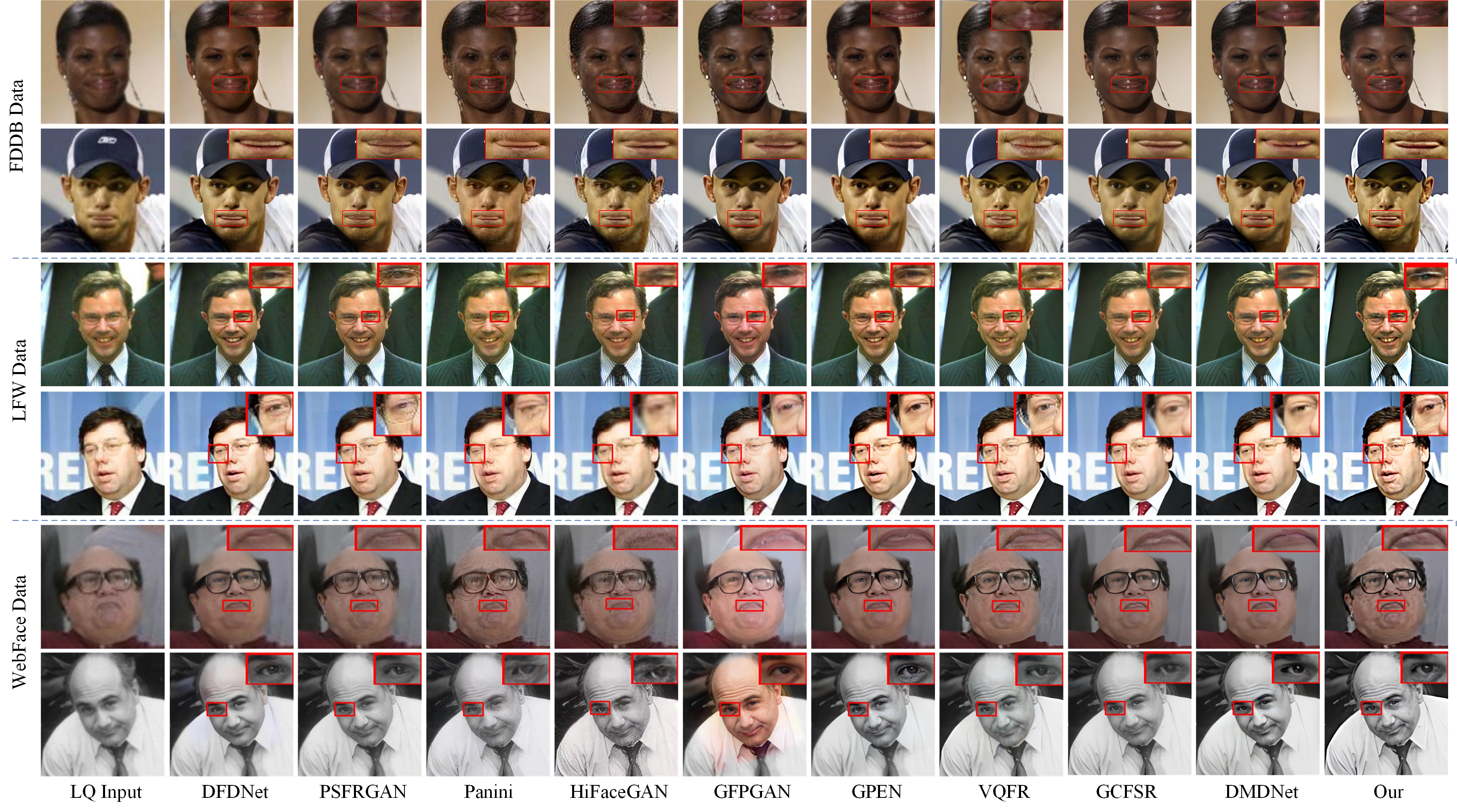}
	\caption{Qualitative comparisons on the \textbf{FDDB Data}, \textbf{LFW Data}, and \textbf{WebFace Data}. \textbf{Zoom in for best view}.}
	\label{fig:BFR_wild}
\end{figure*}
We evaluate our method on the FDDB, LFW, and WedFace Data, which are face images with complex facial poses in real scenes and suffer from multiple complex unknown degradations. To test the generalization ability, we make qualitative and quantitative comparisons with several latest blind face restoration methods, such as DFDNet\cite{li2020DFD}, PSFRGAN\cite{chen2021psfr-gan}, VQFR\cite{gu2022vqfr}, Panini\cite{wang2022panini}, HiFaceGAN\cite{yang2020hifacegan}, GFPGAN\cite{wang2021GFPGAN}, GPEN\cite{yang2021gpen}, DMDNet~\cite{li2022learning}, and GCFSR\cite{he2022gcfsr}. Since there is no GT in the real degraded image, we adopt the non-reference perceptual metrics (e.g., NIQE, IFQA) to test the performance of different methods. For the FDDB data, in order to facilitate qualitative comparison, we utilize the pre-trained RetinaFace face detection model\cite{deng2020retinaface} to cut out a single face with consistent width and height before testing.

The quantitative and qualitative results are presented in Tab.\ref{tab:BFR_wild} and Fig.\ref{fig:BFR_wild}. Among them, our method achieves better IFQA and NIQE values in quantitatively evaluating all datasets. Thanks to local discriminators and an excellent training strategy, our method has more robust generalization when used to restore face images from natural scenes. The restored images have a more vivid and natural texture, more comfortable color information, and higher visual quality. In particular, our results have more realistic details and complete structure in the facial regions. It is also seen that our method can improve the image quality of non-facial regions. In addition, Panini method\cite{wang2022panini} lacks tested results on the FDDB Data. The main reason is that this method requires equal width and height of the input image, but FDDB Data does not meet this condition.

\begin{figure*}[!t] 
	\centering
	\includegraphics[width=1.0\textwidth]{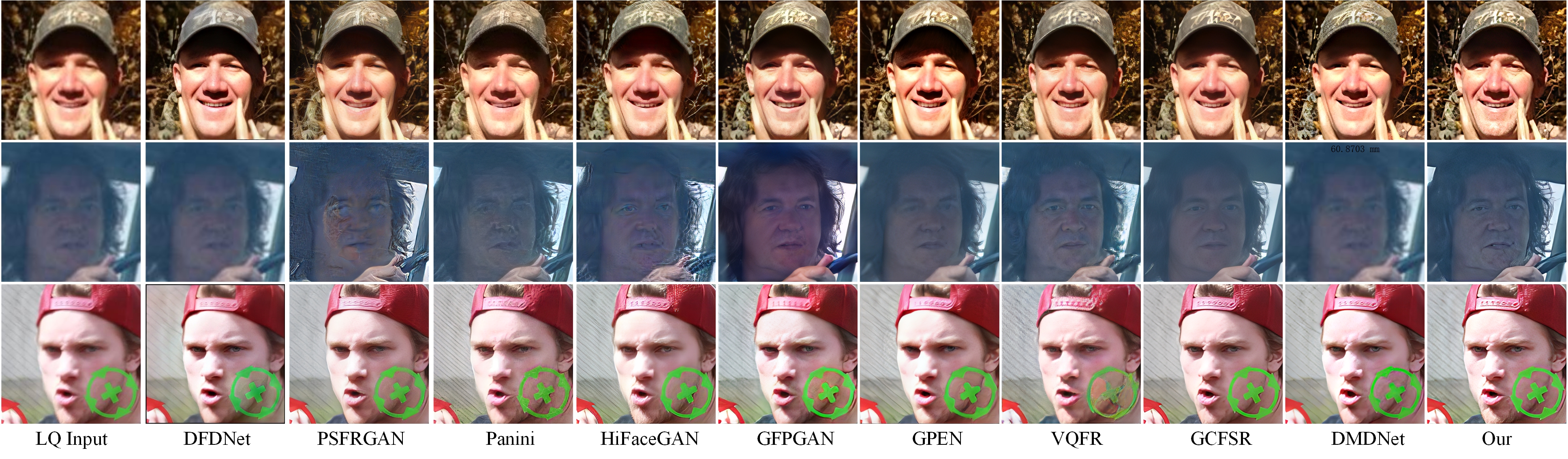}
	\caption{Adaptive restoration results on the \textbf{WFLW Data}. \textbf{Zoom in for best view}.}
	\label{fig:BFR_WFLW}
\end{figure*}

\begin{table}%[]
	%\vspace{-0.3cm}
	%\small
	\centering
    \caption{Comparison of adaptive restoration performance of different methods on the \textbf{WFLW Data}. Only the non-reference perceptual metrics (e.g., NIQE, IFQA) are adopted.}
    \label{tab:BFR_adaptable}
	\scalebox{0.8}{
	%	\hspace{-0.5cm}
    	\begin{tabular}{c|c|c|c|c|c|c}
    		\hline
    		Methods   & Input & DFDNet\cite{li2020DFD} & Panini\cite{wang2022panini} & GCFSR\cite{he2022gcfsr} & VQFR\cite{gu2022vqfr} & DMDNet~\cite{li2022learning} \\ \hline
    		IFQA$\uparrow$    & 0.230   & 0.345 & 0.414    & \textcolor{blue}{\bf 0.449} & 0.416     & \textcolor{red}{\bf 0.452}   \\
            NIQE $\downarrow$   & 10.449 & 5.407 & 4.429    & 5.204 & 4.377     & 5.018    \\ \hline
            Methods & PSFRGAN\cite{chen2021psfr-gan} & HiFaceGAN\cite{yang2020hifacegan} & GFPGAN\cite{wang2021GFPGAN} & GPEN\cite{yang2021gpen} & Our  & \\ \hline  
            IFQA$\uparrow$    & 0.394   & 0.443 & 0.407    & 0.432 & \textcolor{blue}{\bf 0.449}    &   \\
            NIQE $\downarrow$  & \textcolor{blue}{\bf 3.918} & \textcolor{red}{\bf 3.646} & 5.046    & 4.538 & 4.257    &   \\ \hline
	\end{tabular}}
	%\vspace{-0.5cm}
\end{table}
\textit{\textbf{Adaptability in the wild.}} To verify the adaptability of our method in the wild, we have adopted a Wider Facial Landmarks in the Wild (WFLW) dataset containing rich disturbance attributes, such as occlusion, posture, makeup, lighting, blurring, and facial expressions. We randomly select 775 images for testing. Quantitative and qualitative results are shown in Tab.\ref{tab:BFR_adaptable} and Fig.\ref{fig:BFR_WFLW}, respectively. It can be seen that our method has a good adaptive ability for complex situations, especially in occlusion and overexposure, and can still restore realistic and high-fidelity facial images.

\subsection{Ablation Studies}
To better understand the roles of different components and training strategies in our method, we train them separately and apply qualitative and quantitative comparisons to test the performance. The involved relevant variables are as follows: without MMRB layers (w/o MMRB), without Local Discriminators (w/o LocalD), without the FreezeD strategy (w/o FreezeD), without the Face-Preserving loss (w/o FP Loss), and without the Fine-tuning and FreezeD strategy (w/o Ft+FreezeD). As can be seen from the Fig.\ref{fig:ablated study} and Tab.\ref{tab:ablation}, we can find that:
\begin{figure}[!t]
    \centering
    \includegraphics[width=1.0\textwidth]{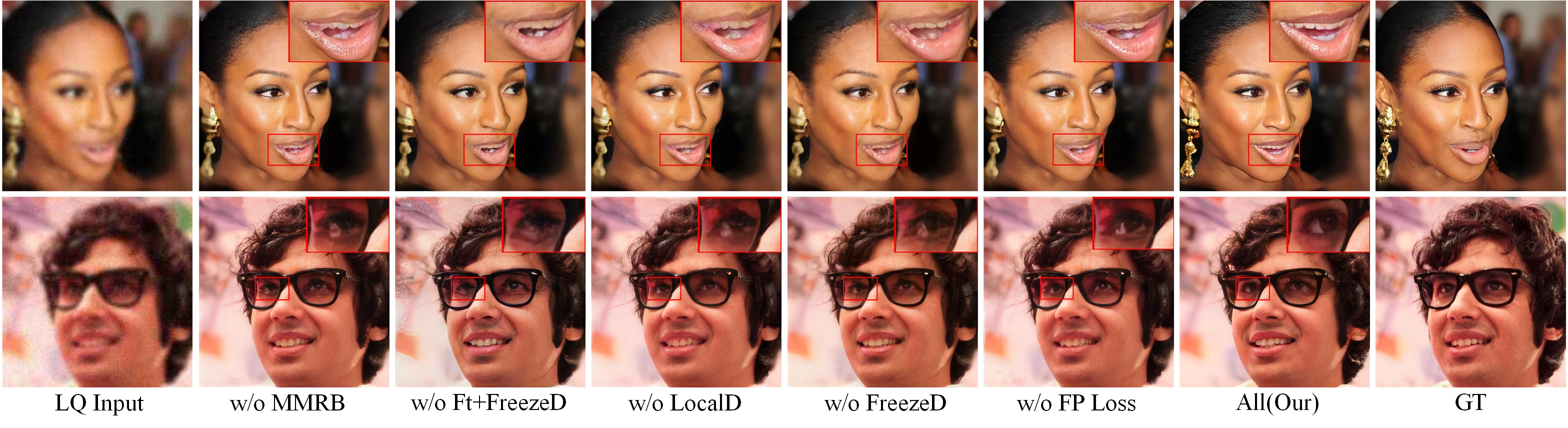}
    \caption{Comparative results of different variables in our model. \textbf{Zoom in for best view}.}
    \label{fig:ablated study}
\end{figure}

\textit{\textbf{MMRB layer.}} If the MMRB layer in the encoder is adopted, it can better maintain the original features of the input image and avoid the excessive illusion of local contents.

\textit{\textbf{Local Discriminator.}}This component significantly affects restoring the structural information of key facial regions, such as the eyes and mouth. If not, the restoration results are unrealistic enough, affecting the visual perception quality.

\textit{\textbf{Face-Preserving loss.}} It significantly influences the restored image's authenticity and fidelity, especially the facial regions' structural information.

\textit{\textbf{FreezeD strategy.}} Using this training strategy can accelerate the training process and better reduce the distribution difference between the restoration results and GT, making the output more realistic and natural. 

\textit{\textbf{Fine-tuning and FreezeD strategy.}} This training strategy can greatly reduce the distribution difference between the restoration results and GT to improve the facial fidelity and the overall quality of the restored image.
\begin{table}
    \centering
	\caption{Quantitative comparisons of different variants on the \textbf{CelebA Data}. Bold \textbf{black} indicates the best performance.}
	\label{tab:ablation}
	\scalebox{1.0}{
		%\hspace{-0.5cm}
    	\begin{tabular}{l|cc|cccc}
    		\hline
    		Components    & PSNR$\uparrow$  & SSIM$\uparrow$  & LPIPS$\downarrow$  & FID$\downarrow$ & NIQE $\downarrow$ &IFQA$\uparrow$ \\ \hline
    		a) w/o MMRB    &  25.977  &  0.699   &  0.163    &  17.027  & 4.682  & 0.367   \\ 
    		b) w/o FreezeD   & 26.016   &  0.701   & 0.164  & 17.812 & 4.868   & 0.343   \\ 
    		c) w/o LocalD  & 26.010    & 0.697     & 0.163  &  16.908  & 4.614 & 0.367  \\ 
    		d) w/o Ft+FreezeD & 25.704   & 0.692   & 0.171  & 19.543  & 4.546 & 0.317  \\ 
            e) w/o FP Loss  & 26.018  & 0.700 & 0.163 & 16.853 & 4.707 & 0.338        \\  \hline
    		\textbf{All (Our)}   & \textbf{26.034}   & \textbf{0.702}    & \textbf{0.162}  & \textbf{16.080}   & \textbf{4.285}  & \textbf{0.370}   \\ \hline
	\end{tabular}}
	%\vspace{-0.3cm}
\end{table}

\section{Running Time}
We compare the running time of state-of-the-art methods \cite{li2020DFD,chen2021psfr-gan,gu2022vqfr,wang2022panini,yang2020hifacegan,wang2021GFPGAN,yang2021gpen,li2022learning,he2022gcfsr} with the proposed model. Here, a single NVIDIA Tesla V100 GPU is mainly employed to evaluate all methods against hundreds of randomly selected 512$\times$512 face images. As shown in Tab.\ref{tab:running_time}, our model can obtain higher IFQA values and better retain facial identity information due to the use of MMRB layers and face-preserving loss. However, MMRB layers also increase the computational complexity of this model.
\begin{table}%[]
	%\vspace{-0.3cm}
	%\small
	\centering
    \caption{Comparison of running time of different methods. “\textbf{-}” means that the result is unavailable.}
    \label{tab:running_time}
	\scalebox{0.8}{
	%	\hspace{-0.5cm}
    	\begin{tabular}{c|c|c|c|c|c|c}
    		\hline
    		Methods    & Input  & DFDNet\cite{li2020DFD} & Panini\cite{wang2022panini} & GCFSR\cite{he2022gcfsr} & VQFR\cite{gu2022vqfr} & DMDNet~\cite{li2022learning} \\ \hline
    		IFQA$\uparrow$ & 0.295 & 0.305 & 0.384   & 0.384 & 0.384     & 0.385   \\
            Time(sec)       & $-$   & 1.056 & 0.088    & 0.029 & 0.177    & \textcolor{red}{\bf 0.021}    \\ \hline
            Methods & PSFRGAN\cite{chen2021psfr-gan} & HiFaceGAN\cite{yang2020hifacegan}  & GPEN\cite{yang2021gpen} & GFPGAN\cite{wang2021GFPGAN} & Our \\ \hline  
            IFQA$\uparrow$    & 0.350   & \textcolor{blue}{\bf 0.393} & 0.352    & 0.382 & \textcolor{red}{\bf 0.421}     & \\
            Time(sec)  & 0.205 & 0.055 & 0.031    & \textcolor{blue}{\bf 0.026}  & 0.109    &\\ \hline
	\end{tabular}}
	%\vspace{-0.5cm}
\end{table}

\section{Conclusion}
Aiming at the problem that complex degraded pose-varied and multi-expression face images are challenging to restore, we meticulously designed a restoration network with a generative facial prior for BFR tasks. To better preserve the original facial features and avoid excessive fantasy, we gradually utilized MMRB layers to extract weak texture features in the input image. Furthermore, we fine-tuned the pre-trained StyleGAN2 model and adopted the FreezeD strategy for the global discriminator model, which can better fit the distribution of diverse face images suitable for natural scenes and improve the quality of the overall images. And especially because eyes and mouth regions are difficult to recover, we exploited different local adversarial losses to constrain our model for these regions. Extensive experiments on synthetic and multiple real-world datasets demonstrate that our model has good generalization capability, could restore complex degraded pose-varied face images, and outperform the latest optimal BFR methods. The restoration results have richer textures, more natural details, and higher facial fidelity. Most importantly, our method can promote the image quality of non-facial regions and restore old photos, film, and television works. Furthermore, it can also be applied in other tasks such as Face Super-resolution and Augmented Reality. 

\section*{Acknowledgment}
This work was partly supported by the Yunnan Major Science and Technology Special Plan Projects (No. 202002AD080001) and the National Natural Science Foundation of China (No. 61771338).

\section*{Declarations}
\textbf{Conflicts of Interest}: The authors declare that they have no existing competing financial interests or personal relationships in this paper.

\textbf{Data Availability Statement}: Data sharing does not apply to this article as no new data were created or analyzed in this study.

%%===========================================================================================%%
%% If you are submitting to one of the Nature Portfolio journals, using the eJP submission   %%
%% system, please include the references within the manuscript file itself. You may do this  %%
%% by copying the reference list from your .bbl file, paste it into the main manuscript .tex %%
%% file, and delete the associated \verb+\bibliography+ commands.                            %%
%%===========================================================================================%%
%\section*{References}
\bibliography{reference}

%\bibliography{sn-bibliography}% common bib file
%% if required, the content of .bbl file can be included here once bbl is generated
%%\input sn-article.bbl

%% Default %%
%%\input sn-sample-bib.tex%

\end{document}